# Offline Handwritten Recognition of Malayalam District Name - A Holistic Approach

Jino P J [#1], Kannan Balakrishnan [*2]

[#] Artificial Intelligence Lab, Department of Computer Applications, CUSAT

Cochin,Kerala,India

[1] jino@cusat.ac.in

[2] mullayilkannan@gmail.com

*Abstract—* **Various machine learning methods for writer independent recognition of Malayalam handwritten district names are discussed in this paper. Data collected from 56 different writers are used for the experiments. The proposed work can be used for the recognition of district in the address written in Malayalam. Different methods for Dimensionality reduction are discussed. Features consider for the recognition are Histogram of Oriented Gradient descriptor, Number of Black Pixels in the upper half and lower half, length of image. Classifiers used in this work are Neural Network, SVM and RandomForest.**

**Keyword- Offline Handwritten Recognition, Machine Learning, Dimensionality Reduction, HOG**

I. INTRODUCTION

Automation of Handwriting recognition is a challenging task because writing styles may vary according to the mental condition, locality, habit, speed etc of writer. Rather than concentrating on the individual letters human beings recognize the words or sentences in a whole manner while reading. In the case of machines we can find mainly three methods for the Handwriting Recognition viz Recognize characters individualy (analytical)[1],whole word(Holistic)[2] and Hybrid Method(Combined Analytical and holistic approach)[3]. In generally template based and feature based methods are used for pattern recognition tasks[4]. Template based feature extraction was popular in earlier days but now handcrafted or machine extracted features along with machine learning techniques are used for the recognition. Spatial domain and transform domain are the two approaches where in the former case it extracts features directly from the image and in later case it transforms image to another space like Fourier or wavelet and extract features from the transformed representation. Most commonly used features in spatial domain methods are topological,statistical,directional and curvature[5].

Major challenge for analytic recognition is the need for a proper segmentation algorithm. Even some times human beings may not be able to segment characters properly, in that case they will recognize it from the context or shape of the word. The proper selection of the feature purely depends on the language domain. In modern research we can find a lot of methods implementing machine extracted features rather than handcrafted features[6][7]. Offline handwritten recognition has several applications like Address Interpretation, Writer Identification, Analyze the progress of a paralyzed patient and with a proper text to speech recognition system it can support visually challenged people. Lack of proper benchmarking database of offline handwritten images is a bottleneck for the researchers in Indian Languages. In Malayalam there is no benchmarking database available.The popular database available in the English language is CEDAR[8], MNIST[9], IAM[10] and CENPARAMI[11].

The paper is organized as follows Section II reviews the literature, Section III explains the details about Malaylam Script and Data collection method for the present work, Section IV describes the Proposed method, Section V discusses Results and Interpretation and Final Section is Conclusion.
.

II. REVIEW OF LITERATURE

In literature we can find several attempts using holistic approach for the recognition of handwritten documents. This section discusses both online and offline handwriting recognition by applying holistic method. Online handwriting recognition of Bangla words achieved 97% accuracy for a lexicon size of 10.This approach uses four feature sets with multiple SVM and the final output is combined[12]. For Devanagary handwritten words Hidden Markov Model based approach with chain code features was in used[13]. Lexicon size for the classification in this work is 100 and shows 80.2 % accuracy.The authors[14] proposed pereceptual feature based classification of handwritten English words using Convolutional Neural Network. They achieved highest accuracy of 93.4% with a sample size of 32,753 for a 23 class problem.In earlier attempts[15] temporal features used for handwritten cursive word recognition with an accuracy of 81% for a lexicon size of 10 with 557 test samples.Directional Features are used for the recognition of Urdu word images is proposed[16] with SVM as the classifier. An accuracy of 97% is achieved with CENPARAMI database with 18,177 samples of 57 lexicons. Hidden Markov Model is used for the historical handwritten document recognition[17].Scalar and profile based features are used for the recognition purpose. Gabor Features along with Number of horizontal lines, vertical lines,dots,+45 slanted lines and -45 slanted lines are used for the recognition of handwritten Tamil words[18]. Classifier used for the recognition purpose is SVM and achieved 91.67% accuracy with 50 classes.

Redundant and least important features should be removed before classification [19].In literature there are several methods for dimensionality reduction of the features by retaining the important one. Harmony search algorithm is used [20]for dimensionality Reduction of features. An accuracy of 90.29 % is shown with a sample size of 1020 handwritten words, where the lexicon size is 20. The above mentioned accuracy is obtained using 48 elliptical features. Stroke orientation distribution features are calculated for the handwritten English legal amount words [21] with PCA reduced features of 3024 to 572. PCA is widely used[12][22][23] for dimensionality reduction.

From the survey it is clear that holistic recognition provides good results to overcome smaller class problems. But in the larger class problems more number of samples are required with carefully designed features.

III. MALAYALAM SCRIPT

India is a linguistically rich country with different popular languages, it includes English as the International language and Hindi is used as the national language. 22 scheduled languages are also used in state level. Malayalam is a south Dravidian classical language used in the state of Kerala and one of the official languages in Puduchey and Lakshadweep. Malayalam consists of two scripts viz oldscript and newscript[24]. The revised form of the Malayalam script consists of 36 consonants and 15 Vowels and a few other characters. Consonants and vowels of Malayalam language are shown in Table I

TABLE I
Malayalam Characters

| Vowels | | | Consonants | | | | |
|---|---|---|---|---|---|---|---|
| അ | ആ | ഇ | ക | ഖ | ഗ | ഘ | ങ |
| ഈ | ഉ | ഊ | ച | ഛ | ജ | ഝ | ഞ |
| ഋ | എ | ഏ | ട | ഠ | ഡ | ഢ | ണ |
| ഐ | ഒ | ഓ | ത | ഥ | ദ | ധ | ന |
| ഔ | അം | അഃ | പ | ഫ | ബ | ഭ | മ |
| | | | യ | ര | ല | വ | ശ |
| | | | ഷ | സ | ഹ | ള | ഴ |
| | | | റ | | | | |

The last two vowels listed in the given Table I are not commonly used in day to day life.

*A. Data Set*

There are fourteen districts in Kerala and so the lexicon size for the present work is 14. Total samples collected from 56 writers are 784.Data is collected from different places and from the different age groups. Samples used for Word recognition is extracted from the forms written by the people in an unconstrained manner, ie, without forcing them to use specific pen, ink colour, line thickness, writing style etc.Sample form for the data collection is shown in Fig I. Some of the writers don't write some words partially or fully and those word samples are neglected. So 48 words from the total samples were neglected and the final sample size for experiments is 736

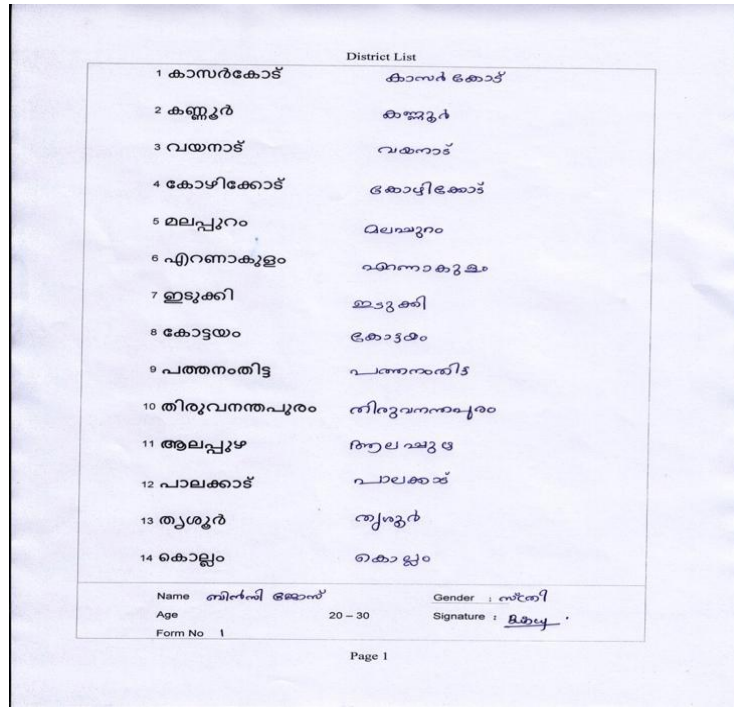

Fig. 1. Sample Form

All the collected forms are digitized using a flat bed scanner with 300 dpi resolution and saved it in tiff format.The information like Age Group,Gender,signature are also collected.

## IV. PROPOSED METHOD

The proposed method has different stages as shown in Fig:2

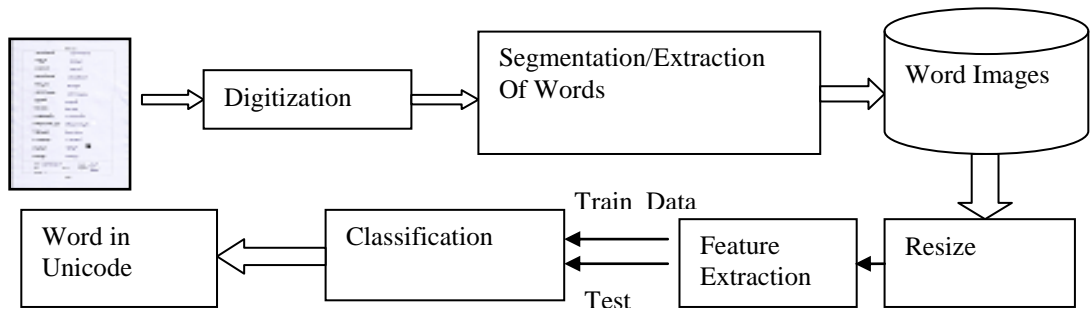

### A. Preprocessing

Extraction of the word images are done through bounding box method[25]. To spot the words exactly,image is dilated and extracted from the original one and converted to grayscale.For the proper recognition of sample word images further processing is required before feeding to the classifier. During the time of scanning if any noise occurs, it should be removed. In our experiments we didn't consider this mainly because samples do not contain much noise. We resize all the images to 64 x 128 pixels by using the bicubic interpolation method[26].

### B. Features

Feature selection is important for the classification task so we select following features, Number of Black Pixels in the upper half and lower half, length of the original image along with HOG (Histogram of Oriented

Gradients)[27]. Later we found that HOG features supercedes these scalar features. So we select HOG for the classification.HOG Feature Descriptor calculates the histograms of directions of oriented gradients. It is highly successful to find out the object shape and thus we select this for our holistic recognition of words.As we mentioned in the preprocessing stage image is converted to 64 rows and 128 columns.Block size, block stride, cell size, number of bins are (16,16),(8,8),(8,8),9 respectively. With this parameters we will get 3780 features. Thus we require a proper dimensionality reduction technique. Original Image and its HOG visualization is shown in Fig II.

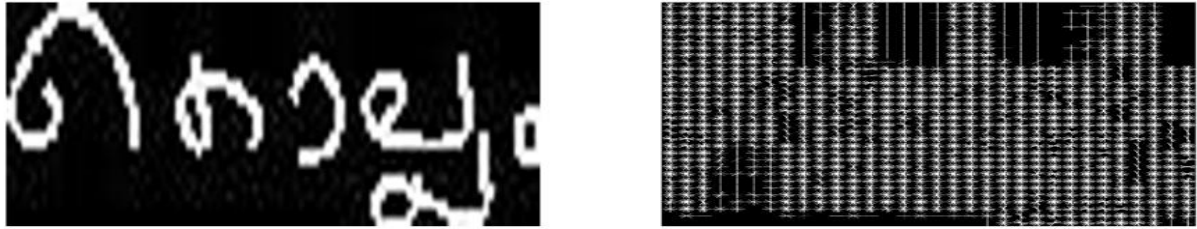

Fig.II: Original Image and its HOG Features Visualization

### C. Dimensionality Reduction

For the reduction of feature dimensions, we consider three methods PCA(Principal Component Analysis), Gaussian Random Projection(GRP) and Sparse Random Projection(SRP)[29]. PCA is a linear dimensionality reduction technique, which uses orthogonal transformation to convert to lower dimensional space. Random Projection is apt for distance based method for approximation. GRP use gaussian random matrix and SRP use sparce random matrix.

### D. Classification and Interpretation

The training and classification are performed through Neural Network,SVM and Random Forest.80 % of the data is given for training and 20%  is given for testing. Same training and testing data is fed through all the classifiers. Best result is achieved with SVM as discussed in Section V. The predicted result is interpreted through Unicode.The class labels and its corresponding Unicode are shown in Table II.

TABLE II
Interpretation class labels with unicode

| District Name | Unicode Mapping In Python |
|---|---|
| കാസർകോട് | u'\u0D15'+u'\u0D3E'+u'\u0D38'+u'\u0D7C'+u'\u0D15'+u'\u0D4B'+u'\u0D1F'+u'\u0D4D' |
| കണ്ണൂർ | u'\u0D15'+u'\u0D23'+u'\u0D4D'+u'\u0D23'+u'\u0D42'+u'\u0D7C' |
| വയനാട് | u'\u0D35'+u'\u0D2F'+u'\u0D28'+u'\u0D3E'+u'\u0D1F'+u'\u0D4D' |
| കോഴിക്കോട് | u'\u0D15'+u'\u0D4B'+u'\u0D34'+u'\u0D3F'+u'\u0D15'+u'\u0D4D'+u'\u0D15'+u'\u0D4B'+u'\u0D1F'+u'\u0D4D' |
| മലപ്പുറം | u'\u0D2E'+u'\u0D32'+u'\u0D2A'+u'\u0D4D'+u'\u0D2A'+u'\u0D41'+u'\u0D31'+u'\u0D02' |
| എറണാകുളം | u'\u0D0E'+u'\u0D31'+u'\u0D23'+u'\u0D3E'+u'\u0D15'+u'\u0D41' +u'\u0D33'+u'\u0D02' |
| ഇടുക്കി | u'\u0D07'+u'\u0D1F'+u'\u0D41'+u'\u0D15'+u'\u0D4D'+u'\u0D15'+u'\u0D3F' |
| കോട്ടയം | u'\u0D15'+u'\u0D4B'+u'\u0D1F'+u'\u0D4D'+u'\u0D1F'+u'\u0D2F'+u'\u0D02' |
| പത്തനംതിട്ട | u'\u0D2A'+u'\u0D24'+u'\u0D4D'+u'\u0D24'+u'\u0D28'+u'\u0D02'+u'\u0D24'+u'\u0D3F'+u'\u0D1F'+u'\u0D4D'+u'\u0D1F' |
| തിരുവനന്തപുരം | u'\u0D24'+u'\u0D3F'+u'\u0D30'+u'\u0D41'+u'\u0D35'+u'\u0D28'+u'\u0D28'+u'\u0D4D'+u'\u0D24'+u'\u0D2A'+u'\u0D41'+u'\u0D30'+u'\u0D02' |
| ആലപ്പുഴ | u'\u0D06'+u'\u0D32'+u'\u0D2A'+u'\u0D4D'+u'\u0D2A'+u'\u0D42'+u'\u0D34' |
| പാലക്കാട് | u'\u0D2A'+u'\u0D3E'+u'\u0D32'+u'\u0D15'+u'\u0D4D'+u'\u0D15'+u'\u0D3E'+u'\u0D1F'+u'\u0D4D' |
| തൃശൂർ | u'\u0D24'+u'\u0D43'+u'\u0D36'+u'\u0D42'+u'\u0D7C' |
| കൊല്ലം | u'\u0D15'+u'\u0D4A'+u'\u0D32'+u'\u0D4D'+u'\u0D32'+u'\u0D02' |

## V. RESULTS AND INTERPRETATION

All the experiments are performed in a System with Intel Core i7-4770CPU@3.40GHz processor with 8GB RAM. Neural network and SVM are implemented in Python, Random Forest(RF) is implemented in R in Linux platform.

### A. Neural Network

Classical Neural networks always perform well for character recognition. The reduced Feature set is fed to the MLP(Multi Layer Perceptron) Classifier and result analyzed. MLP is an algorithm that learns in a supervised manner[28]. It learns a function $f: \mathbb{R}^m \rightarrow \mathbb{R}^o, m \in \{50, 100, 316, 733, 1523\}, o = 14$. The model consists of a 3 Layer Architecteure, Input Layer with neurons equals to the number of features and hidden layer with 100 neurons. Optimizer is Stochastic Gradient Descent. Values from the input layer are transformed by each neuron in the hidden layer with a weighted linear summation followed by ReLU activation function[30] defined as $g(x) = \max(0, a)$. ReLU always provide maximum between 0 and pre activation, so it is always non negative. Loss function is categorical cross entropy. Softmax function used in the output layer for the prediction and it consists of 14 neurons. Precision, Recall and f1-Score after testing is shown in Table III.

TABLE III
Result Analysis of Neural Network with 100 neurons

| Label | Precision | Recall | f1-score | Support |
|---|---|---|---|---|
| 1 | 0.75 | 1.00 | 0.86 | 6 |
| 2 | 0.80 | 0.92 | 0.86 | 13 |
| 3 | 0.88 | 0.88 | 0.88 | 8 |
| 4 | 1.00 | 0.92 | 0.96 | 13 |
| 5 | 1.00 | 0.93 | 0.96 | 14 |
| 6 | 1.00 | 0.92 | 0.96 | 12 |
| 7 | 0.91 | 0.83 | 0.87 | 12 |
| 8 | 1.00 | 1.00 | 1.00 | 7 |
| 9 | 0.92 | 0.86 | 0.89 | 14 |
| 10 | 0.90 | 1.00 | 0.95 | 9 |
| 11 | 0.91 | 1.00 | 0.95 | 10 |
| 12 | 0.90 | 1.00 | 0.95 | 9 |
| 13 | 1.00 | 0.88 | 0.93 | 8 |
| 14 | 0.83 | 0.77 | 0.80 | 13 |
| **avg /** | **0.92** | **0.91** | **0.91** | **148** |

The maximum number of neurons in hidden layer implemented is 350, after that there is no further improvement in the result. Table IV shows the result and Fig III analyse the performance in terms of accuracy.

TABLE IV
Comparison with changes in Hidden Layer

| No. Of Neuron in Hidden Layer | Accuracy |
|---|---|
| 50 | 89.18 |
| 100 | 91.21 |
| 150 | 93.24 |
| 350 | 94.59 |

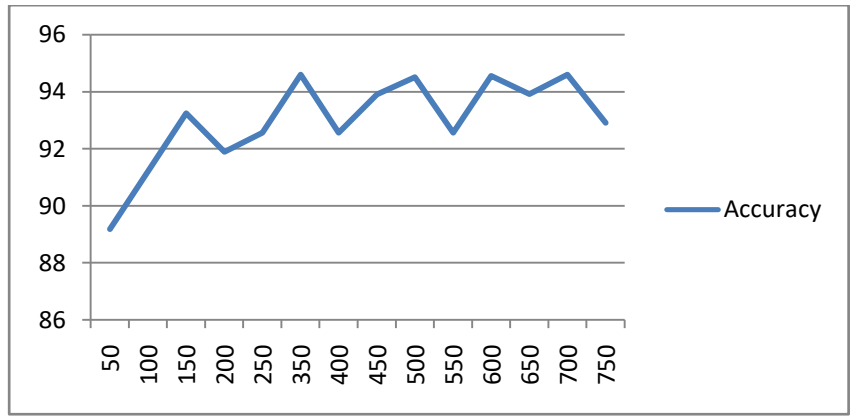

Fig. III : Performance with increase in hidden layer neurons

## B. Support Vector Machne (SVM) Based Classifier

SVM 's are easy and efficient method for classification. Data fed to the SVM should be in the following format.Features, F= {$f_1,f_2,f_3,........,f_n$}  Label,L={$l_1,l_2,l_3,.........,l_n$} Where the value of n should be same for both Features and Labels Fi can be a list of features extracted from the handwritten images where i=1,2,3,4,....n.  In our experiments n is 736 and Features extracted initially for an image is 3780. SVM uses four basic kernels like Linear,Polynomial,RBF(Radial Bais Function) and sigmoid. These kernels have different parameters and in our experiment RBF kernel[31] with Gridsearch method for parameter selection for the classification.Result analysis is shown in Table V.

TABLE V
Result Analysis with SVM

| Label | Precision | Recall | f1-score | Support |
|---|---|---|---|---|
| 1 | 0.86 | 1.00 | 0.92 | 6 |
| 2 | 1.00 | 1.00 | 1.00 | 13 |
| 3 | 1.00 | 1.00 | 1.00 | 8 |
| 4 | 1.00 | 0.92 | 0.96 | 13 |
| 5 | 1.00 | 1.00 | 1.00 | 14 |
| 6 | 0.92 | 0.92 | 0.92 | 12 |
| 7 | 1.00 | 1.00 | 1.00 | 12 |
| 8 | 1.00 | 1.00 | 1.00 | 7 |
| 9 | 1.00 | 0.93 | 0.96 | 14 |
| 10 | 0.90 | 1.00 | 0.95 | 9 |
| 11 | 0.91 | 1.00 | 0.95 | 10 |
| 12 | 1.00 | 1.00 | 1.00 | 9 |
| 13 | 1.00 | 0.88 | 0.93 | 8 |
| 14 | 0.92 | 0.92 | 0.92 | 13 |
| avg / total | 0.97 | 0.97 | 0.97 | 148 |

## C. Random Forest(RF)

RF is a tree structured classifier [32] and its contribution to the medical imaging is notable. Number of trees considered for classification is 50,100,2000 respectively. After 50/100/2000 trees is produced,they vote for the most popular class.These procedures are called random forest. Random Forest algorithm will create ensemble of trees $\{T_b\}_1^m$ , where m is the number of trees. Consider  one of the experiment performed with 100 features with 100 trees, here 100 trees will form and each node in a tree will use a subset of features selected randomly, by default the number will be $\sqrt{100}$ = 10.All the trees are trained independently and parallely. Average the trees output will predict the probability of a given sample belongs to certain class shown in equation given below.

$$P(C|v) = \frac{1}{100}\sum_{t=1}^{100} P_t(C|v)$$

The performance with the classifier is depicted in fig.IV

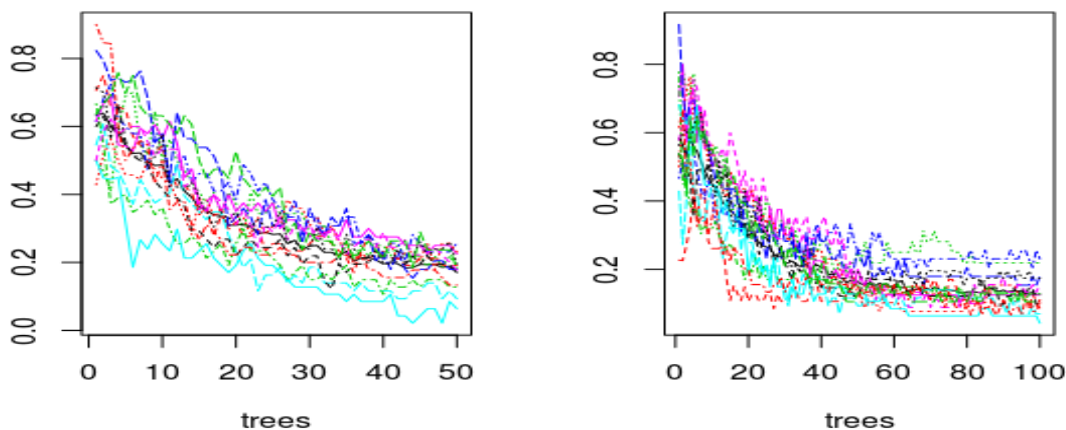

Fig IV: Error Analysis with 50 and 100 trees

From the graph it is clear that after 50 trees the performance is only improving slightly.Result analysis with different trees are shown in table IV

TABLE VI
Comparison of Results

| No.of Trees | Accuracy |
|---|---|
| 50 | 87.16 |
| 100 | 87.83 |
| 2000 | 90 |

Reduced dimensions using the techniques mentioned in section IV and corresponding result with above mentioned classifiers are shown in Table VII.

TABLE VII
Result Analysis With Different Dimensionality Reduction Methods

| | | | Features=50 | Features=100 | |
|---|---|---|---|---|---|
| PCA | Neural Network | | 93.24% | 94.59% | |
| | **SVM** | | **95%** | **97 %** | |
| | RF | 50 trees | 85.81% | 87.16% | |
| | | 100trees | 88.51% | 87.83% | |
| | | | Features=316 | Features=733 | Features=1523 |
| GRP | Neural Network | | 87.8% | 91.89% | 91.21 |
| | SVM | | 89% | 94% | 95% |
| | RF | 50 trees | 67.56% | 78.37% | 75.67% |
| | | 100trees | 76.3% | 83.78% | 85.13% |
| SRP | Neural Network | | 91.21 | 88.51 | 89.86 |
| | SVM | | 91% | 92% | 94% |
| | RF | 50 trees | 74.32% | 79.05% | 77.70% |
| | | 100trees | 79.72% | 82.43% | 84.45% |

In terms of accuracy PCA outperforms over Random Subspace Projection Method with all the classifiers.SVM got comparatively good accuracy of 97 % with 100 Features.

## VI. CONCLUSION

Holistic method of handwritten word recognition is implemented with the help of three popular machine learning classifiers like Neural Network,SVM, and RandomForest. SVM with RBF kernel provide 97% Result with PCA as the dimensionality reduction approach. Lexicon was less mainly because the total number of districts in Kerala is 14. The work can be extended to more lexicon size. Also we can consider handcrafted features along with machine extracted features for the recognition.

ACKNOWLEDGMENT

Authors acknowledge all the writers those who contribute for the creation of dataset. First author is greateful to CUSAT for funding through University Junior Research Fellowship Programme.

## AUTHOR PROFILE

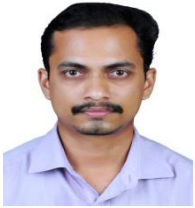

**Jino P J**, received his M. Sc. Degree in Computer Science from BharathiarUniversity, Coimbathore, Thamilnad and MBA from IGNOU.He is currently pursuing the Ph.D degree at Cochin University of Science and Technology, Cochin, India. His research interests are in handwriting recognition, document image analysis, pattern recognition and machine learning . He has published papers in international journals and conference proceedings.

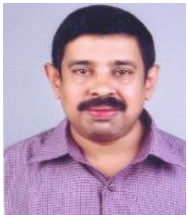

**Dr. Kannan Balakrishnan**, born in 1960, received his M. Sc and M.Phil degrees in Mathematics fromUniversity of Kerala, India, M. Tech degree in Computer and Information Science from Cochin University of Science & Technology, Cochin, India and Ph. D in Futures Studies from University of Kerala, India in 1982, 1983, 1988 and 2006 respectively. He is currently working with Cochin University of Science & Technology, Cochin, India, as an Associate Professor, Head in the Department of Computer Applications. He has visited Netherlands as part of a MHRD project on Computer Networks.Also he was the co investigator of Indo-Slovenian joint research project by Department of Science and Technology, Government of India. He has published several papers in international journals and national and international conference proceedings. His present areas of interest are graph Algorithms, Intelligent systems, Image processing, and language computing. He is a reviewer of American Mathematical Reviews.